# EDIT: Enhancing Vision Transformers by Mitigating Attention Sink through an Encoder-Decoder Architecture

**Wenfeng Feng [a,*], Hongxiang Wang[a] , Jianlong Wang[a,*], Xin Zhang[a],Jingjing Zhao[a] , Yueyue Liang[a], Duokui Han[a],Xiang Chen[a]**

School of Computer Science and Technology, Henan Polytechnic University, Jiaozuo, 454003, China

fengwf@hpu.edu.cn (Wenfeng Feng ), 212409020022@home.hpu.edu.cn (Hongxiang Wang), wangjianlong24@hpu.edu.cn (Jianlong Wang ),

**Abstract**: In this paper, we propose EDIT (Encoder-Decoder Image Transformer), a novel architecture designed to mitigate the attention sink phenomenon observed in Vision Transformer (ViT) models. Attention sink occurs when an excessive amount of attention is allocated to the [CLS] token, distorting the model's ability to effectively process image patches. To address this, we introduce a layer-aligned encoder-decoder architecture, where the encoder utilizes self-attention to process image patches, while the decoder uses crossattention to focus on the [CLS] token. Unlike traditional encoder-decoder framework, where the decoder depends solely on high-level encoder representations, EDIT allows the decoder to extract information starting from low-level features, progressively refining the representation layer by layer. EDIT is naturally interpretable demonstrated through sequential attention.

## 1. Introduction

Transformer, introduced by Vaswani et al. [1], utilize self-attention and cross-attention mechanisms to extract intrinsic features from text data. Transformer includes both an encoder and a decoder, with the encoder extracting relevant information from input data and the decoder generating outputs based on this representation. Transformer and its improvements have achieved significant success in natural language processing (NLP) tasks [1, 2, 3, 4, 5].

Improvements to the transformer architecture have led to two main branches: First, decoder-only models, like the GPT series [3, 4, 5], are suited for generative tasks. Second, encoder-only models, such as BERT [2], are tailored for classification tasks.

The success of Transformers in the field of natural language processing (NLP) has inspired researchers to explore their applications in computer vision (CV), ultimately leading to the development of Vision Transformers (ViT) [6]. ViTs are widely applied in tasks such as image classification, object detection and segmentation, and video analysis, and have become a reasonable alternative to convolutional architectures. As a general-purpose architecture, Transformers can learn convolutional operations as well as long-range interactions through the attention mechanism. ViT directly processes sequences of image patches for classification tasks and can be regarded as an adaptation of the BERT model (an encoder-only model) for computer vision. It transforms image patches and positional embeddings into input sequences and introduces a special [CLS] token at the beginning of the sequence to aggregate information for classification purposes. In contrast, convolutional networks [7] inherently provide built-in translation invariance, eliminating the need to explicitly learn such priors during training. As a result, hybrid architectures that incorporate convolution layers tend to converge faster than vanilla Transformers [8].

The ViT model employs self-attention to learn relationships among these tokens, thus, resulting in three types of attention: First, self-attention among input tokens to representations image features. Second, cross-attention from the [CLS] token to input tokens to assess feature importance for classification. Third, attention from input tokens to the [CLS] token, which maybe unnecessary or even detrimental. The integration of these attention types can introduce irrelevant information and increase model complexity, and may compel the model to optimize conflicting objectives: guiding self-attention among patches while summarizing information for the linear classifier.

One piece of evidence supporting our point of view is the attention sink phenomenon we found in ViT. The phenomenon of attention sink was first found in transformer-based language models [9] refers to the tendency for a disproportionate amount of attention to be allocated to the initial tokens in a sequence, regardless of their semantic relevance. This occurs due to the SoftMax function in attention mechanisms, which requires attention scores to sum to one. Consequently, when there is no strong match for the current query among earlier tokens, excess attention is often directed towards these initial tokens, making them act as 'sinks' for unused attention values.

Similar to the attention sink phenomenon observed in transformer-based language models, we have identified an analogous phenomenon in ViT models, as illustrated in fig. 1. In these models, a disproportionate amount of attention is allocated to the [CLS] token compared to other input tokens. This bias towards the [CLS] token arises from the mixture of input tokens and the [CLS] token within ViT models. The blending of these tokens can lead to conflicting objectives during attention processing, ultimately affecting the model's ability to effectively represent image features for classification tasks.

To illustrate the phenomenon of attention sink, we downloaded two ViT-based image classification models[10]pre-trained on ImageNet-1k: deit3_small_patch16_224.fb_in1k and deit3_base_patch16_224.fb_in1k, both obtained from the HuggingFace model repository. We then extracted the attention maps from all 12 layers of these models and computed the amount of attention allocated to the [CLS] token, alongside the mean attention value directed towards the other input tokens. These results were then visualized in fig. 1 to effectively demonstrate the attention distribution across the layers and highlight the disproportionate focus on the [CLS] token.

This disproportionate focus on the [CLS] token suggests an over-centralization of attention resources on the classification token, potentially at the expense of richer information integration from other input tokens. This phenomenon could limit the model's ability to effectively leverage information from the entire input sequence, underscoring the need for explicitly separating the [CLS] token from the input tokens to mitigate attention sink.

In order to address this mixture of input tokens and the [CLS] token, we refer to the original encoder-decoder architecture, and propose to explicitly separate the [CLS] token from the input tokens. Our new architecture use the original self-attention layers to capture the attention between input tokens, and introduce new cross-attention layers to extract the attention of the [CLS] token to the input tokens. We argue that this explicit separation avoids the contradiction and complexity of the traditional [CLS] token in ViT, and improves performance.

In summary, our main contributions are as follows:

- The article introduces a novel encoder-decoder image transformer architecture (EDIT), designed to address the attention sink phenomenon present in Vision Transformers (ViT). By explicitly separating input tokens from the [CLS] token and employing cross-attention mechanisms with layer alignment, this architecture enhances information processing efficiency.
- The layer-wise attention maps generated by the EDIT architecture intuitively illustrate the process by which the model progressively focuses on critical features, thereby significantly improving the interpretability of Vision Transformers.

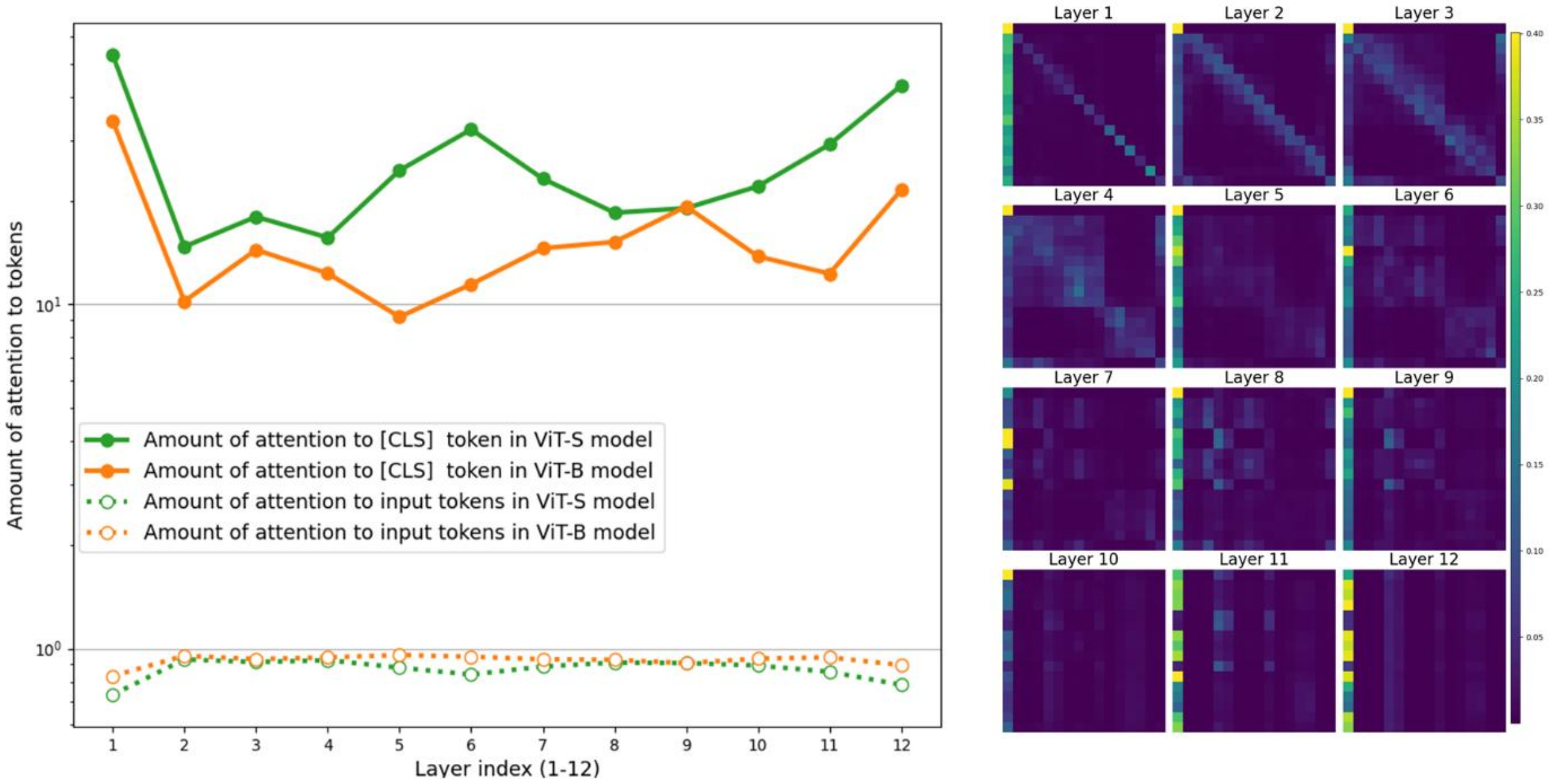

Fig. 1. The phenomenon of attention sink in ViT models. (a) The [CLS] token consistently receives much higher (more than 10 times) attention values, while attention directed toward other tokens remains relatively low and stable throughout the layers. (b) Visualization of the attention scores across all 12 layers in ViT-S model. We visualize the starting 16 × 16 submatrix of the 197 × 197 attention maps of each layer. We observe that the model consistently focuses on the [CLS] token across all layers.

## 2. Related work

### *2.1. Transformer for Vision*

In ViT, an image $\mathbf{x} \in \mathbb{R}^{h \times w \times c}$ is split into a sequence of flattened patches $\mathbf{x} \in \mathbb{R}^{n \times (p^2 \cdot c)}$ where $c$ is the number of channels, $(h, w)$ is the resolution of the original image, $(p, p)$ is the resolution of each image patch, and $n = hw/p^2$ is the number of patches. Each flattened patch is then linearly projected to a patch embedding to produce a sequence of patch embeddings $\mathbf{x}_p \in \mathbb{R}^{n \times d}$ where $d$ is the embedded dimension size (or model dimension size). To capture positional information, learnable position embeddings $\mathbf{pos} \in \mathbb{R}^{n \times d}$ are added to the sequence of patch embeddings to get the resulting sequence of patch and position embeddings $\mathbf{p}_0 = \mathbf{x}_p + \mathbf{pos}$. After that, a learnable class embedding $\mathbf{c}_0 \in \mathbb{R}^{d}$ is concatenated to the sequence of patch and position embedding to produce the input sequence of tokens $\mathbf{z}_0 = \mathrm{concat}(\mathbf{p}_0, \mathbf{c}_0) \in \mathbb{R}^{(n+1) \times d}$ for the ViT backbone.

The ViT backbone composed of $l$ layers is applied to the sequence of tokens $\mathbf{z}_0$ to generate a sequence of encodings $\mathbf{z}_l \in \mathbb{R}^{(n+1) \times d}$. A ViT layer consists of a multi-head self-attention (MSA) block followed by a feed-forward network (FFN) block with layer norm (LN) applied before every block and residual connections added after every block:

$$\mathbf{ai{-}1} = \mathbf{MSA}(\mathbf{LN}(\mathbf{zi{-}1})) + \mathbf{zi{-}1}, \tag{1}$$

$$\mathbf{zi} = \mathbf{FFN}(\mathbf{LN}(\mathbf{ai{-}1})) + \mathbf{ai{-}1}, \tag{2}$$

where $i \in \{1, \ldots, l\}$.

The self-attention mechanism is composed of three point-wise linear layers mapping tokens to three intermediate representations, queries $\mathbf{q} \in \mathbb{R}^{(n+1) \times d}$, keys $\mathbf{k} \in \mathbb{R}^{(n+1) \times d}$ and values $\mathbf{v} \in \mathbb{R}^{(n+1) \times d}$. The simplified computational flowchart is shown in fig. 2a. Self-attention is then computed as follows:

$$\mathbf{MSA}(\mathbf{q}, \mathbf{k}, \mathbf{v}) = \mathrm{softmax}\left(\mathbf{qk}T \sqrt{d}\right) \mathbf{v}. \tag{3}$$

In the cross-attention mechanism, two input sequences of the same dimension are asymmetrically combined, with one sequence serving as queries $\mathbf{q} \in \mathbb{R}^{(n)\times d}$ , and the other as keys $\mathbf{k} \in \mathbb{R}^{(n)\times d}$ , and values $\mathbf{v} \in \mathbb{R}^{(n)\times d}$ inputs. The simplified computational flowchart is shown in fig. 2b. Cross-attention is computed as follows:

$$\mathbf{CA}(\mathbf{q}, \mathbf{k}, \mathbf{v}) = \text{softmax}\ (\ \mathbf{qk}T \sqrt{\ d}\ )\ \mathbf{v}. \tag{4}$$

A FFN block is applied after the MSA block. It consists of two linear transformations and a nonlinear activation function within them, and can be denoted as the following function:

$$\mathbf{FFN}(\mathbf{a}) = \mathbf{w2} \cdot \sigma(\mathbf{w1} \cdot \mathbf{a}), \tag{5}$$

where $\mathbf{w1}$ and $\mathbf{w2}$ are two learnable matrices of the two linear transformations, $\sigma$ represents the nonlinear activation function, such as GELU, and $\mathbf{a}$ is the output of the MSA block.

The output of ViT backbone $\mathbf{z}_l$ is feed into the ViT head, who first splits the generated encodings $\mathbf{z}_l$ into two partitions, the patch and position encodings $\mathbf{p}_l \in \mathbb{R}^{n\times d}$ and the class encoding $\mathbf{c}_l \in \mathbb{R}^{d}$ , then extracts the class encoding. After that, a linear transformation is applied to the class encoding $\mathbf{c}_l$ to produce a logits vector. A softmax layer is then used to transform the logits into class probabilities.

The use of special tokens to extend transformer sequences became prominent with BERT [2]. Typically, these tokens serve to introduce new information (e.g., [SEP] in BERT), allocate additional computation to the input (e.g., tape tokens in AdaTape, [11]), or aggregate specific outputs for tasks such as classification ([CLS] in BERT and ViT [6]), generative learning ([MASK] in BERT and BEiT [12]), or object detection (e.g., object queries in DETR [13], and ViDT [14]). In multimodal systems, latent token arrays are used to merge information before decoding, as demonstrated in Perceivers [15, 16]. Memory Transformer [17] introduced memory tokens to enhance sequence processing, boosting translation tasks. Subsequent studies, such as [18], extended this to solve complex sequence operations like copyreverse tasks. [19] applied this concept in the vision domain for task-specific fine-tuning, although these tokens were found to lack generalization across tasks.

Researchers have recognized the problem of mixing input tokens with the [CLS]. In a later paper [20], Beyer, one of the authors of the ViT model, have suggested that using 1D Global Average Pooling in place of the [CLS] token could enhance performance. Touvron et al. [21] proposed to only allow the [CLS] token to attend to the patch tokens, rather than the opposite.

Visualizing attention maps from the [CLS] token to patch tokens has been widely adopted since its introduction in DINO [22]. DINO demonstrated that its attention maps were free from artifacts that plagued earlier Vision Transformers. Building on this, other studies have proposed techniques to refine attention maps, such as modifying optimization strategies [23], guiding attention scores to focus on salient image regions [24], altering transformer architectures [25], or incorporating learnable pooling mechanisms to refine [CLS] token representations [26].

### *2.2. Encoder-Decoder architecture*

The original transformer architecture, introduced by Vaswani et al. [1], was designed for sequence-to-sequence tasks such as language translation, specially English-to-French and English-to-German. It consists of two main components: an encoder and a decoder. The encoder is responsible for processing and understanding the input text (source language). It converts the input sequence into a continuous representation, or embedding, which captures the relevant information from the source text. This embedding is then passed to the decoder for further processing. The decoder takes the continuous representation from the encoder along with the previously generated translated tokens, and generates the output sequence (target language). It uses the encoder's output and the autoregressive property of the model to sequentially generate the translated text, predicting one token at a time based on the context provided by the encoder's embeddings and prior decoded tokens. Together, the encoder and decoder form the basis of the transformer architecture, which enables effective handling of complex sequence-to-sequence tasks by leveraging self-attention and cross-attention mechanisms for capturing dependencies across tokens in the input and output sequences.

Recent advancements in encoder-decoder transformer models, such as BART [27] and T5 [28, 29], have significantly enhanced performance in various NLP tasks. These models combine innovative techniques, pre-training objectives, and architectural modifications to leverage the strengths of both the encoder and the decoder components. Encoder-decoder models are particularly effective for tasks involving input and output sequences with varying lengths or structures. They excel in applications where it is crucial to model the relationships between elements in both sequences. Common use cases include text translation, summarization, question answering, and text generation. By capturing contextual information in the input sequence and generating a coherent output, these models have become central to a variety of NLP tasks.

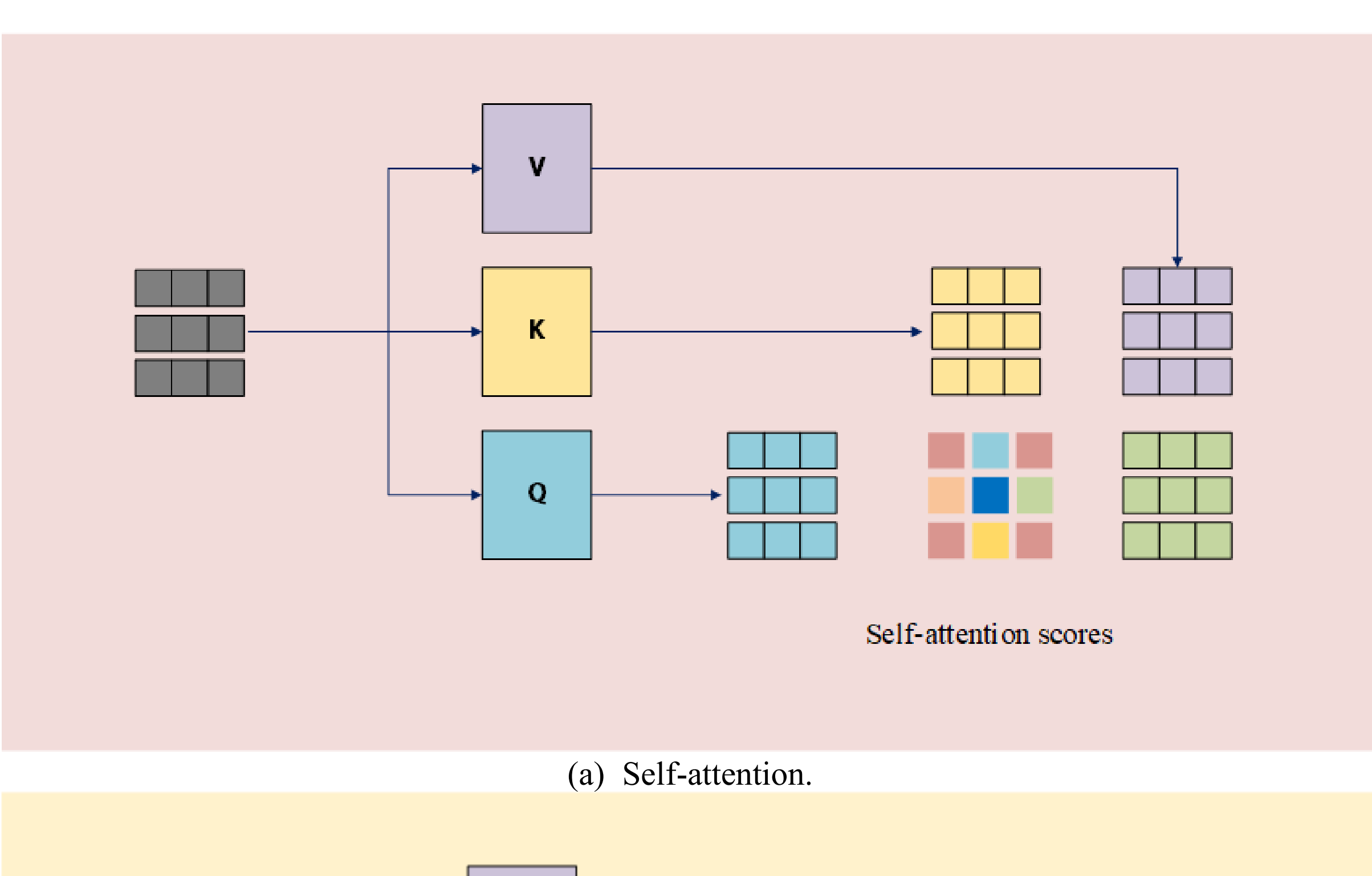

(a) Self-attention.

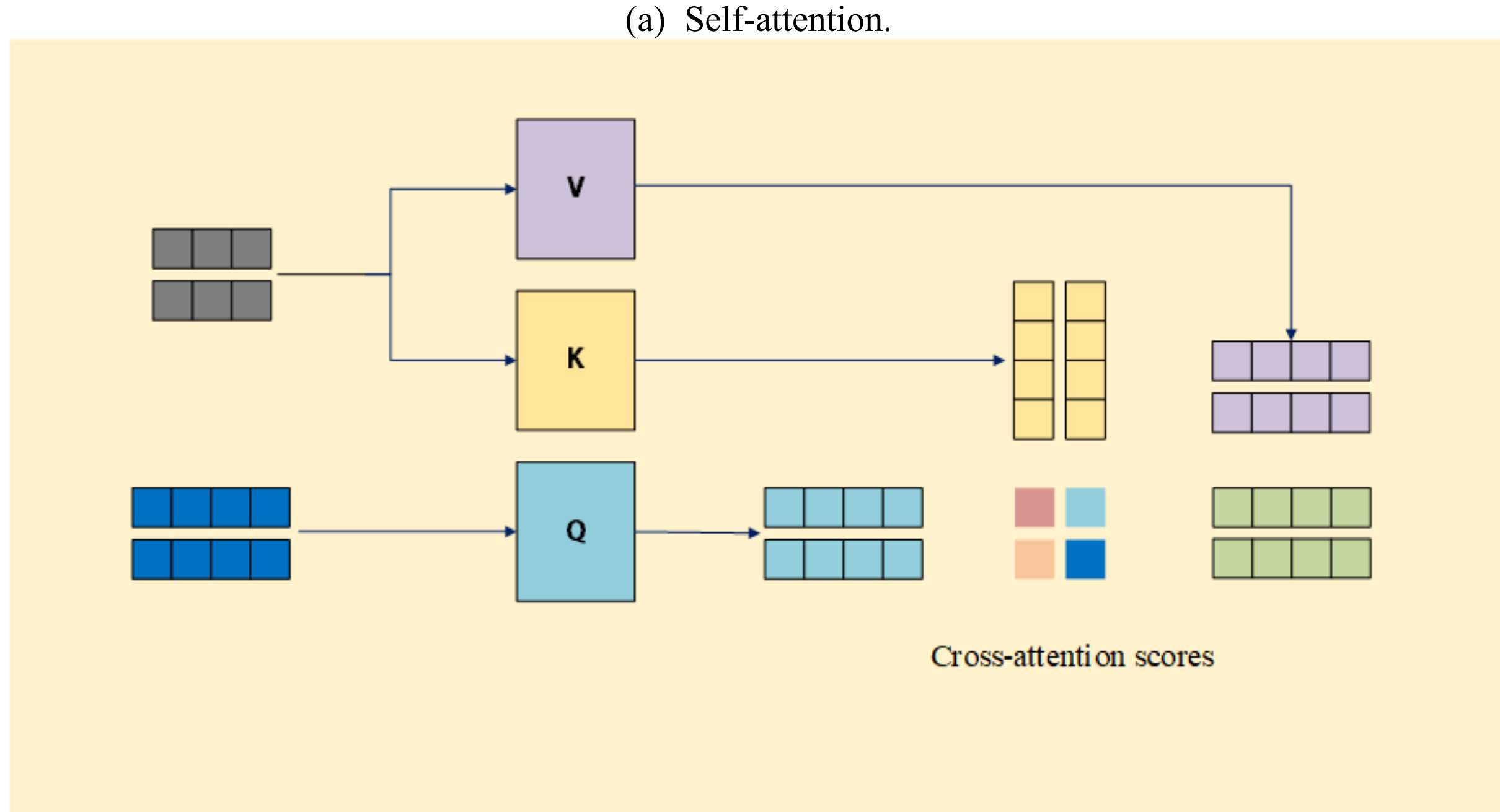

(b) Cross-attention.

Fig. 2. The Attention Calculation Process.

## 3. Method

Our proposed architecture is based on a novel encoder-decoder framework, where the encoder and decoder are aligned layer by layer. Unlike the traditional encoder-decoder architecture, where the decoder typically relies on the high-level representations from the encoder, our architecture maintains a layer-wise alignment, enabling both components to interact more effectively at each stage of processing.

### *3.1. Layer-by-Layer Alignment*

In the traditional encoder-decoder architecture, the encoder processes the input tokens (source tokens) and generates a set of hidden representations, progressing from low-level to high-level features. These representations are then passed to the decoder, which generates target token representations, typically starting from the highest-level representations of the input.

However, this design raises a few natural questions: Why should the low-level representations of target tokens depend solely on the highest-level representations of the source tokens? Why not allow each decoder layer to attend to the corresponding layer of the encoder, preserving more detailed information at each stage?

In contrast to this, our architecture avoids this high-level dependency by aligning the encoder and decoder layers. Specially, each layer of the decoder is aligned with its corresponding encoder layer, ensuring that the encoder and decoder operate in tandem at the same level of abstraction.

In our design, the encoder consists of standard self-attention layers that process the input tokens (image patches and position embeddings). The decoder, on the other hand, generates hidden representations for the [CLS] token by attending to the corresponding encoder layer and the previous decoder layer via a cross-attention mechanism.

The $i$-th layer of the encoder processes the $i$-th level of the input tokens, while the $i$-th layer of the decoder generates the target token representations based on both the source information from the $i$-th encoder layer and the previously generated target tokens from the $(i-1)$-th decoder layer. This design enables the decoder to extract information starting from low-level features, progressively refining the representation at each layer, rather than relying only on the high-level encoded information from the final encoder layer.

This layer-wise alignment has been shown to improve performance in tasks such as Neural Machine Translation (NMT), where precise alignment of encoder-decoder layers has been found to boost translation quality [30].

### *3.2. Model Architecture (EDIT)*

The whole model architecture is illustrated in the upper part of fig. 3, which mainly includes two parts:

- Tokenizer Block: The input image patches are first transformed into a sequence of tokens via the Tokenizer block.
- Backbone: The backbone of the model consists of $N$ identical layers, where each layer takes two parts: one from the encoder (tokens of image patches and position embeddings) and the other from the decoder ([CLS] token). These two inputs are processed in parallel, with outputs from the encoder and decoder feeding into the next layer.

At the final layer, the output of the [CLS] token is passed through a linear transformation and a SoftMax layer to produce output probabilities.

Each layer of the model consists of two aligned sub-layers illustrated in the lower part of fig. 3:

- Encoder Layer: This layer is identical to the standard ViT layer, which includes a multi-head self-attention mechanism followed by a fully connected feed-forward network. Both sub-layers are preceded by layer normalization, with residual connections around each of them.
- Decoder Layer: The decoder layer contains a single-head cross-attention mechanism, which enables the decoder to attend to both the encoder's output and the previously generated decoder output. This cross-attention mechanism is followed by a layer normalization and a residual connection.

It is important to note that the decoder layers in our model share weights, as indicated by the dashed outline in the diagram in fig. 3. This weight-sharing strategy helps reduce model complexity and ensures consistent information flow across layers.

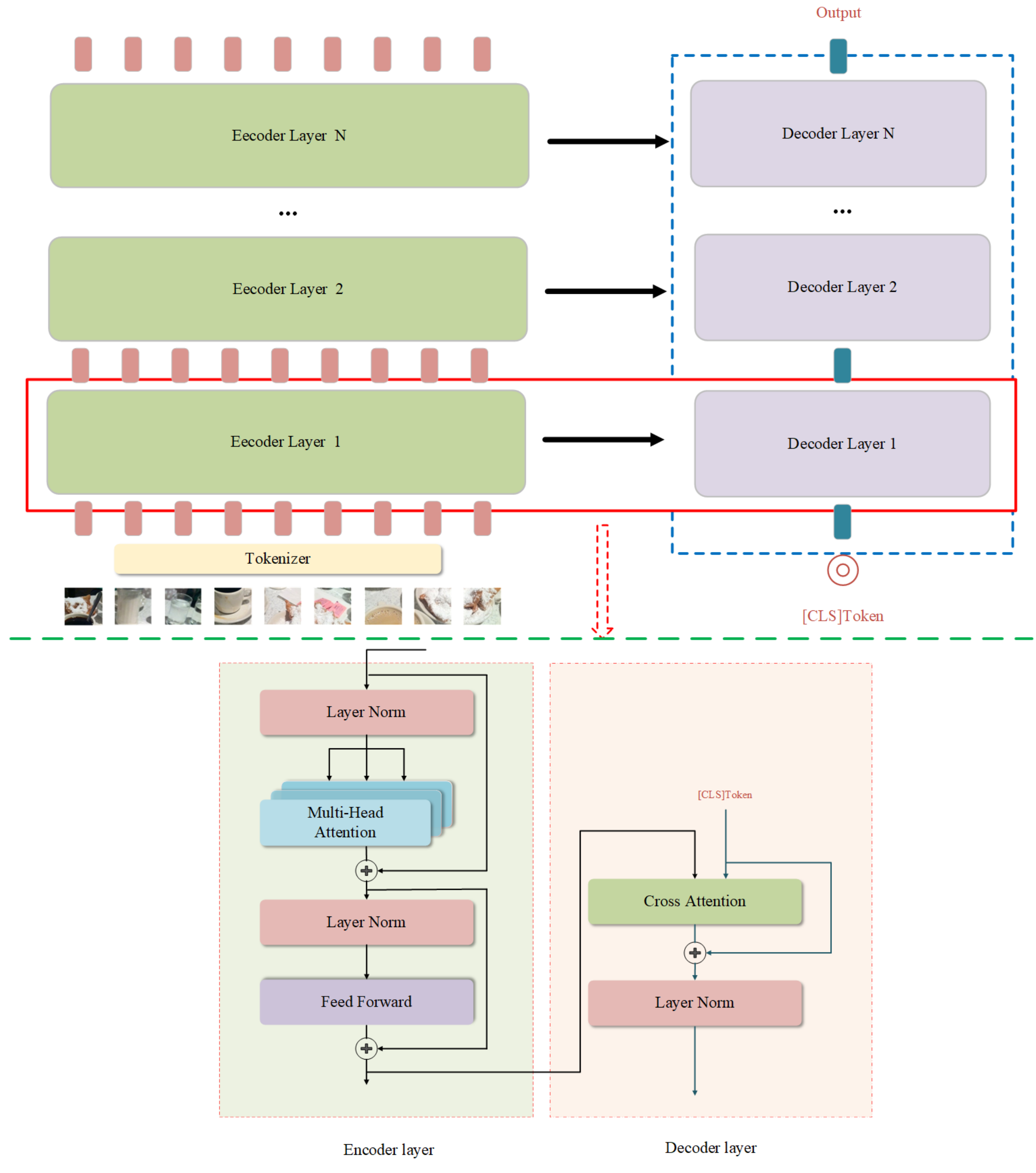

Fig. 3. The architectures of EDIT model.

By aligning the encoder and decoder layers and incorporating cross-attention at each stage, our architecture improves the efficiency and effectiveness of information processing in ViT-based image classification tasks. The design ensures that the decoder has access to more fine-grained information, which helps in better feature extraction and more accurate classification, leading to improved overall model performance.

## 4. Experiments

### *4.1. computational resources*

Due to the limitation of having only NVIDIA 3090/4090 GPU cards, which do not support distributed training across nodes, we are restricted in training large models on extensive datasets. Therefore, we limited our largest model to the EDIT-Base configuration and constrained our largest dataset to ImageNet-21K with an image size of 224x224. This restriction ensures that our experiments remain feasible within the computational resources available.

### *4.2. Datasets*

For pre-training, we use two large-scale image datasets. First is ImageNet-1k, i.e. the ILSVRC-2012 ImageNet dataset [31], containing 1,000 classes and approximately 1.3 million images. Second is a selected subset of the ImageNet-21k dataset [32, 31]. The original ImageNet-21k contains around 14 million images and 21,000 object categories. We first clean the dataset by removing infrequent classes with fewer than 500 images, resulting in 11,582,723 images across 10450 classes. We then allocate 50 images per class to create a standardized validation split.

For transfer learning evaluation, we use 5 popular computer vision datasets:CIFAR-10, CIFAR-100 [33], Oxford Flowers-102 [34], Describable Textures Dataset (DTD) [35], and Caltech-256 [36].

For semantic segmentation, we use the ADE20k dataset [37]. This dataset consists of 20k training images and 5k validation images labeled with more than 150 categories.

### *4.3. Model Variants and Training Procedures*

We base the configurations of EDIT models on those used in DeiT3 [38] and DeiT [10], as detailed in table 1. The EDIT-Small and EDIT-Base models are adopted directly from DeiT3-Small and DeiT3-Base, while EDIT-Tiny integrates the original DeiT-Tiny model due to the absence of a DeiT3-Tiny configuration. As described in section 3.2, the EDIT models are constructed by adding iterative, weight-sharing decoder layers to the DeiT3 architecture. Given computational constraints, we limit our configurations to models no larger than DeiT3-Base.

Table 1. Variants of our EDIT architecture

| Model | Hidden size | Heads | Params | FLOPs |
|---|---|---|---|---|
| EDIT-Tiny | 192 | 3 | 5.82M | 1.42G |
| EDIT-Small | 384 | 6 | 22.30M | 5.28G |
| EDIT-Base | 768 | 12 | 87.76M | 20.27G |

We first train our models on ImageNet-1k and ImageNet-21k datasets, then transfer them to five downstream datasets: CIFAR-10, CIFAR-100, Flowers-102, DTD and Caltech-256. table 2 indicates the hyper-parameters that we use by default for all our experiments.

Our training procedures follow those of DeiT3 with a few adjustments: First, for the EDIT-Tiny model, the stochastic depth rate is reduced from 0.05 to zero, as suggested by the DeiT-Tiny configuration due to the lack of a DeiT3-Tiny model. Second, we only performed EDIT-Small, EDIT-Base, DeiT3-Small, and DeiT3-Base trained on ImageNet-21k dataset for 240 epochs. Last, due to computational constraints, we use gradient accumulation in the timm library [39] to reduce GPU memory requirements for the DeiT3-Base and EDIT-Base models. This adjustment results in a slight decrease in accuracy on ImageNet-1k, from 83.8 to 83.7, in our experiments.

For transfer learning, we reduce learning rate and batch size, adjust weight decay and stochastic depth rate, etc. to prevent overfitting on smaller datasets.

Table 2.Hyper-parameters of our models when trained on ImageNet-1k and ImageNet-21k, and transferred to CIFAR-10, CIFAR-100, Flowers-102, DTD and Caltech-256 datasets.

| Methods | ImageNet-21k | ImageNet-1k | |
|---|---|---|---|
| | Pretrain | Pretrain | Transfer |
| Epochs | 240 | 800 | 400 |
| Batch size | 2048 | 2048 | 768 |
| Optimizer | LAMB | LAMB | LAMB |
| Learning rate | $3.10^{-3}$ | $3.10^{-3}$ | $1.10^{-4}$ |
| Learning rate decay | cosine | cosine | cosine |
| Weight decay | 0.02 | 0.02 | 0.01 |
| Warmup epochs | 5 | 5 | 5 |
| Label smoothing $\varepsilon$ | 0.1 | ✗ | 0.1 |
| Stoch. Depth | 0.05 | 0.05 | 0.1 |
| Repeated Aug | ✗ | ✓ | ✗ |
| Gradient Clip | 1.0 | 1.0 | ✗ |
| H. flip | ✓ | ✓ | ✗ |
| RRC | ✗ | ✓ | ✗ |
| 3 Augment | ✓ | ✓ | ✓ |
| LayerScale | ✓ | ✓ | ✓ |
| Mixup alpha | ✗ | 0.8 | ✗ |
| Cutmix alpha | 1.0 | 1.0 | 1.0 |
| ColorJitter | 0.3 | 0.3 | ✗ |
| Eval crop ratio | 1.0 | 1.0 | 1.0 |
| Loss | CE | BCE | CE |

### *4.4. Comparison to ViT models*

The comparison of our models to ViT models is shown in table 3 and fig. 4.

As shown in table 3, the EDIT models have a slight increase in parameter count parameter count and FLOPs compared to their DeiT3 counterparts, with EDIT-Tiny, EDIT-Small, and EDIT-Base showing a 2%, 1%, and 1% increase in parameters, a 12%, 14%, and 15% increase in FLOPs, respectively. This increase in computational cost is expected due to the iterative and weight-sharing decoder layers added in EDIT models.

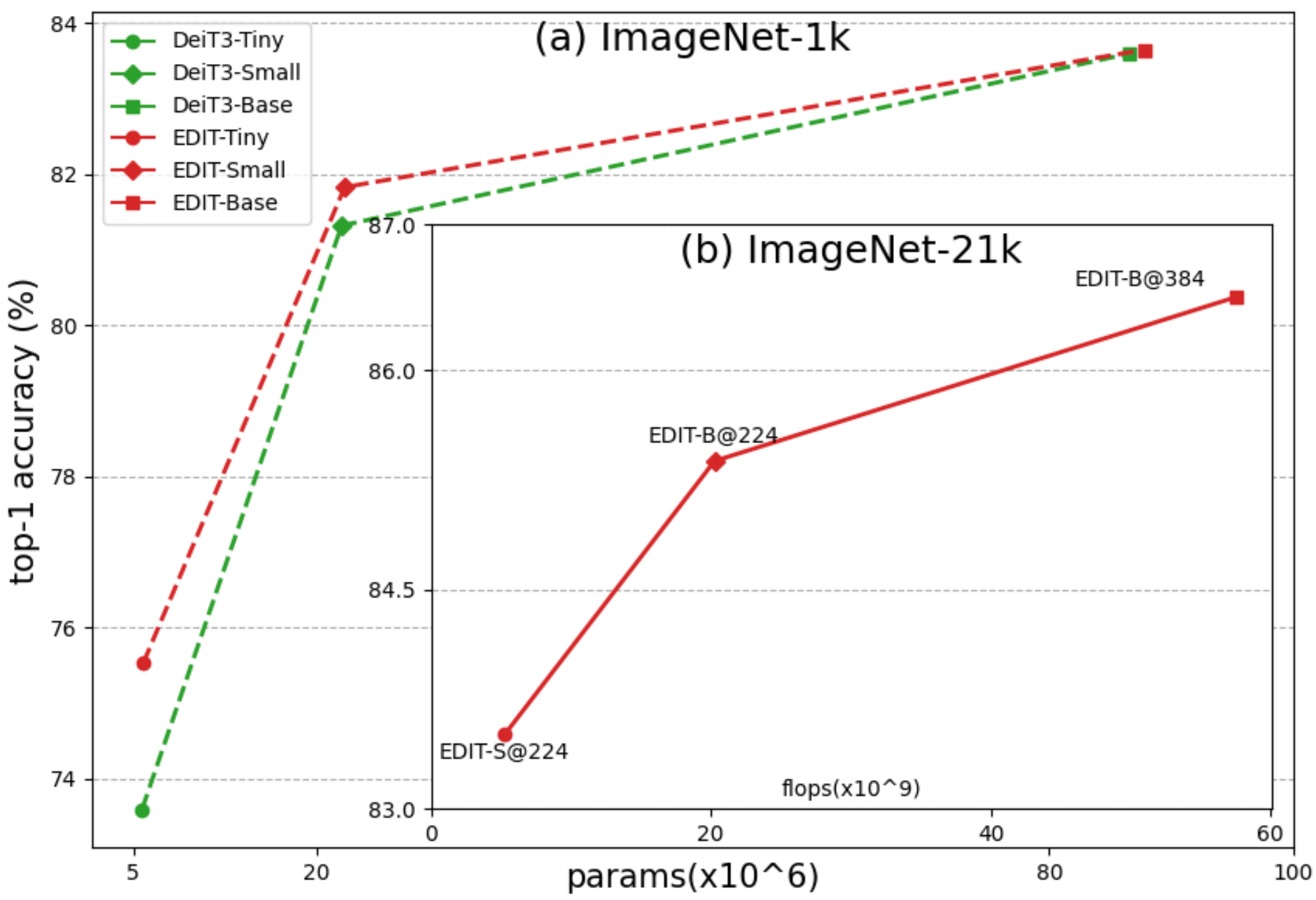

Fig. 4. (a) Comparison of top-1 accuracy between EDIT model and DeiT3 models when trained on ImageNet-1k and (b) pre-trained on ImageNet-21k at 224 × 224 and fine-tuned on ImageNet-1k at resolution 224 × 224 or 384 × 384.

On ImageNet-1k, we observe that the accuracy improvement of EDIT over DeiT3 diminishes as model size increase from Tiny to Small and then to Base. This trend suggests that for a relatively smaller dataset like ImageNet-1k, larger models may encounter diminishing returns due to insufficient data to fully leverage the added capacity. As model size increases without a corresponding increase in data volume or complexity, the model's capacity may become underutilized, resulting in smaller gains.

Conversely, on the larger ImageNet-21k dataset, the accuracy improvement of EDIT over DeiT3 grows as model size scales from Small to Base. This observation aligns with the deep learning scaling law [40, 41], as larger models can more effectively leverage the additional data and complexity provided by the ImageNet-21k dataset. The increase data volume and diversity allow larger models to better generalize, thus enhancing the benefits of the EDIT architecture as model size increase.

An additional insight from these observation is that a mismatch between model size and data size can lead to suboptimal performance in either direction. When a small model is paired with a large dataset, underfitting can occur, as the model lacks sufficient capacity to capture the full complexity of the data. Conversely, a large model paired with a small dataset can lead to overfitting, as the model may focus on dataset-specific noise rather than learning robust, generalizable patterns. This underlines the importance of selecting a model size that aligns with the scale and diversity of the training data to achieve balanced and effective learning.

In summary, these results illustrate that while the EDIT models bring slight increase in computational cost, they generally offer accuracy improvements. The scaling law in vision models is evident here: larger datasets effectively utilize the capacity of larger models, while smaller datasets may lead to overfitting in high-capacity architectures.

### *4.5. Interpreting EDIT models thought attention maps*

The interpretability of vision transformers [42, 43, 44, 45] is crucial in understanding how these models process visual data to make decisions, shedding light on their inner mechanisms. The goal is to discern how ViTs allocate attention across different parts of an image as they progress through each layer, gradually focusing on the most relevant features that contribute to the final classification or recognition.

We observe that our EDIT models are naturally interpretable by the sequential attention maps of each layer. The interpretability is likely caused by the encoder-decoder architecture with iteratively weight-sharing decoder layers that explicitly separates the [CLS] token and input tokens. We illustrate the interpretability by two examples as shown in fig. 5.

Table 3. Comparison of parameters, throughput, and accuracy of different models on ImageNet-1K. We reproduced the results of DeiT3, with the results of ViT and DeiT referenced from [6] and [10], respectively.

| | (a) Regular ImageNet-1K trained models | | | | (b) ImageNet-21K pre-trained models | | | |
|---|---|---|---|---|---|---|---|---|
| Model | image size | Params | FLOPs | ImageNet top1(%) | image size | Params | FLOPs | ImageNet top1(%) |
| ViT-B/16 | $384^2$ | 86M | 55.4G | 77.9 | $384^2$ | 86M | 55.4G | 84.0 |
| ViT-L/16 | $384^2$ | 307M | 190.7G | 76.5 | $384^2$ | 307M | 190.7G | 85.4 |
| DeiT-T | $224^2$ | 5M | 1.1G | 72.2 | - | - | - | - |
| DeiT-S | $224^2$ | 22M | 4.3G | 79.8 | - | - | - | - |
| DeiT-B | $224^2$ | 86M | 16.9G | 81.8 | - | - | - | - |
| DeiT3-T | $224^2$ | 6M | 1.3G | 73.6 | - | - | - | - |
| DeiT3-S | $224^2$ | 22M | 4.6G | 81.3 | $224^2$ | 22M | 4.6G | 83.1 |
| DeiT3-S | $384^2$ | - | - | - | $384^2$ | 22M | 12.5G | 84.6 |
| DeiT3-B | $224^2$ | 87M | 17.6G | 83.6 | $224^2$ | 87M | 17.6G | 85.1 |
| DeiT3-B | $384^2$ | - | - | - | $384^2$ | 87M | 49.4G | 86.1 |
| EDIT-T | $224^2$ | 6M | 1.4G | 75.5 | - | - | - | - |
| EDIT-S | $224^2$ | 22M | 5.3G | 81.8 | $224^2$ | 22M | 5.3G | 83.5 |
| EDIT-S | $384^2$ | - | - | - | $384^2$ | 22M | 14.5G | 84.8 |
| EDIT-B | $224^2$ | 88M | 20.3G | 83.7 | $224^2$ | 88M | 20.3G | 85.4 |
| EDIT-B | $384^2$ | - | - | - | $384^2$ | 88M | 57.5G | 86.3 |

First example is a beautiful dessert. The main part of the dessert is a triangular piece of chocolate cake. On top of the cake, there is a dollop of white cream, and the cream is decorated with small chocolate balls. The chocolate cake is placed on a layer of dark-brown chocolate sauce, which forms a circular base on the white plate.

fig. 5a illustrate how our EDIT model learns to identify the dessert in the picture. In the first few layers, the model primarily focuses on the broader context of the image, capturing the edges of the white plate and the outer boundary of the dark-brown chocolate sauce. This stage reflects the model's early processing, where it seeks to understand the overall scene layout and identify significant boundaries. As the model progresses through the layers, it begins to narrow its focus, reducing attention on the outer plate edge. This refinement indicates that the model has started to prioritize regions that are more relevant to the dessert itself, recognizing that the plate's outer edge is less essential for identification. In the later layers, the model increasingly concentrates its attention on the central part of the dessert, particularly the triangular chocolate piece and the cream with chocolate balls. By the final layers, attention to extraneous parts of the image is minimized, and focus is directed to the key visual elements crucial for identifying the dessert.

Second example is a lovely white kitten with a pair of bright blue eyes in an indoor environment with some furniture and items in the background.

fig. 5b illustrate how our EDIT model learns to identify the kitten in the picture. The model begins by observing general shapes, including background items and the kitten's silhouette. With more layers, it gradually narrows down to the kitten's distinguishing features, particularly the face and bright blue eyes, which are essential for identifying the kitten accurately.

This progression reveal the inner working mechanism of EDIT: starting from a broad view and progressively narrowing focus to critical features. Through this stepwise attention shift across layers, the EDIT model demonstrates a refined interpretability, effectively filtering out non-essential parts of the image.

### *4.6. Downstream tasks*

*4.6.1. Transfer learning.* In order to evaluate the quality of the EDIT models learned through our training procedure we evaluated them with five transfer learning tasks. We focus on the performance of EDIT-Base model pre-trained on ImageNet-1k at resolution 224× 224. The transfer learning procedures are also indicated in table 2. Our results are presented in table 4. The results indicate that EDIT-Base outperforms DeiT3-Base across all tasks, albeit by small margins. An exception is the task of classification on CIFAR-100 dataset where a 0.5% accuracy improvement is achieved. The results highlight

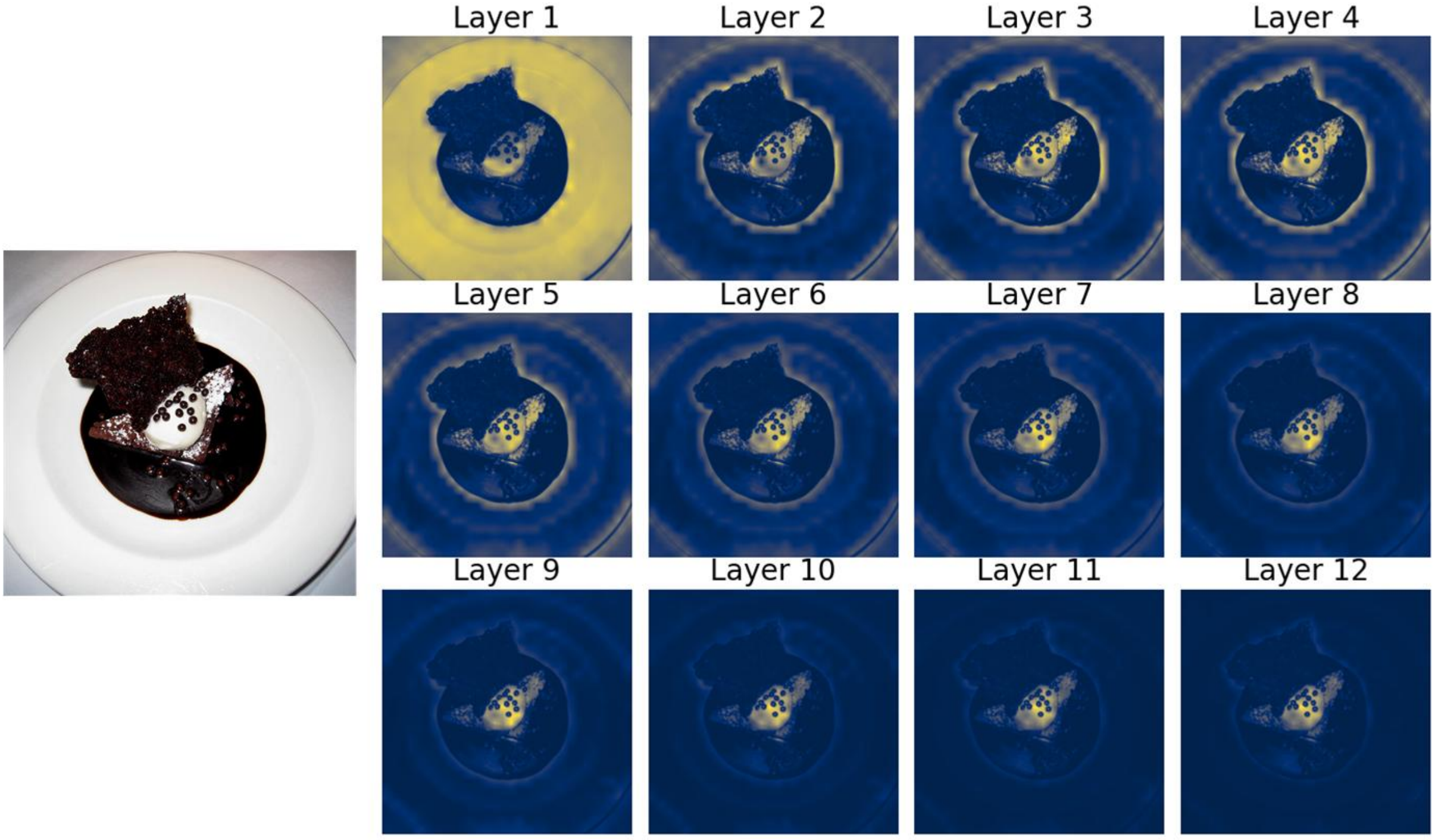

(a) The attention maps of 12 layers in EDIT model when deal with the dessert picture.

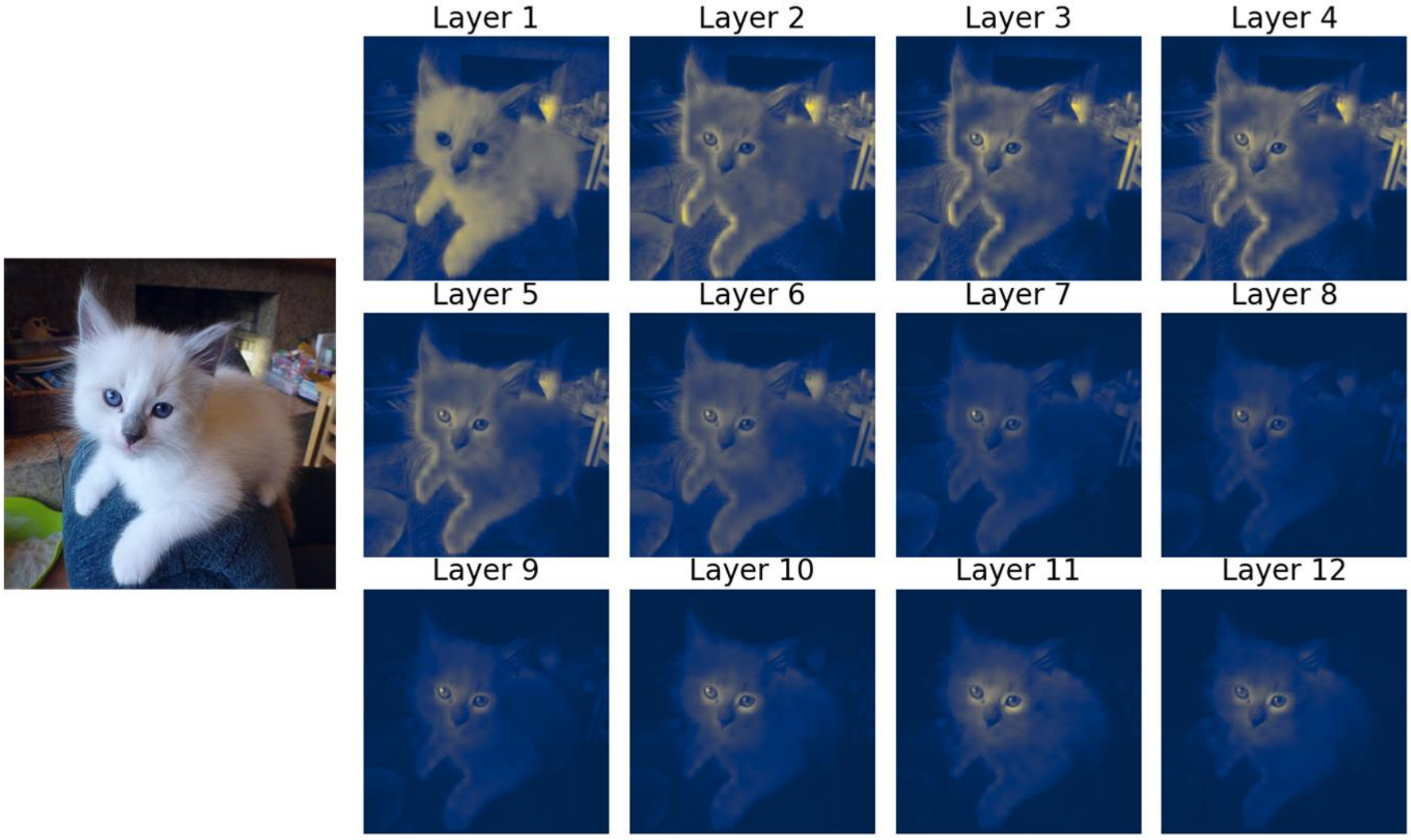

(b) The attention maps of 12 layers in EDIT model when deal with the cat picture.

Fig. 5. Visualization of the attention maps of all 12 layers in EDIT model.

the effectiveness of EDIT-Base in transferring knowledge across diverse datasets, and suggest that the EDIT model may better capture transferable features than DeiT3 model.

Table 4. We compare Transformers based models on different transfer learning tasks with ImageNet-1k pre-training.

| Model | CIFAR-100 | CIFAR-100 | Flowers-102 | DTD | Caltech-256 |
|---|---|---|---|---|---|
| ViT-B/16 | 98.1 | 87.1 | 89.5 | - | - |
| DeiT-B | 99.1 | 90.8 | 98.4 | - | - |
| DeiT3-B | 99.3 | 92.5 | 98.6 | 81.5 | 94.1 |
| EDIT-B | **99.3** | **93.0** | **98.9** | 81.4 | **94.2** |

*4.6.2. Semantic segmentation*.We evaluated the performance of our model on the ADE20K dataset through semantic segmentation experiments. Since our study focuses on Transformer-based models, Segmenter was selected as the baseline method. For training, we adopted the same schedule as Segmenter, using Mask Transformer for 64 epochs of iterative optimization. As the original Segmenter paper did not include results for the DeiT3 model on semantic segmentation tasks, we supplemented our experiments with performance metrics for DeiT3. table 5 reports the comparative results of mean IoU performance across models. The experimental results indicate that under comparable computational costs, EDIT-S achieves an improvement of 1.9 mIoU over ViT-S and 2.0 mIoU over DeiT3-S in single-scale evaluations. Furthermore, in multiscale evaluations, EDIT achieves the best performance with 51.5 mIoU, surpassing other ViT-based models. These findings suggest that EDIT exhibits superior capability in capturing contextual information while maintaining a similar model size and computational budget.

Table 5. Results of semantic segmentation on the ADE20K.

| ADE20K Method | Backbone | Params | Single scale mIoU | Multi-scale mIoU |
|---|---|---|---|---|
| Seg-T-Mask | ViT | 7M | 38.1 | 38.8 |
| Seg-S-Mask | ViT | 27M | 45.3 | 46.9 |
| Seg-B-Mask | ViT | 106M | 48.5 | 50.0 |
| Seg-B-Mask | DeiT | 87M | 47.1 | 48.1 |
| Seg-S-Mask | DeiT3 | 26M | 45.2 | 46.7 |
| Seg-B-Mask | DeiT3 | 103M | 49.0 | 50.8 |
| Seg-S-Mask | EDIT | 27M | 47.2 | 47.8 |
| Seg-B-Mask | EDIT | 105M | 49.9 | 51.5 |

## 5. Conclusion

In this work, we introduced EDIT, an encoder-decoder image transformer designed to address the attention sink phenomenon prevalent in ViT models. By aligning encoder and decoder layers and incorporating layer-specific cross-attention, EDIT avoids the traditional [CLS] token integration issues that contribute to suboptimal attention distribution. Our experiments demonstrate that EDIT consistently outperforms DeiT3 counterparts across various tasks, particularly on larger datasets where model capacity can be effectively utilized. The alignment with deep learning scaling laws further supports EDIT's suitability for scalable visual tasks. Additionally, the sequential attention maps show that EDIT progressively narrows its focus on relevant features, enhancing model interpretability. Future work could extend EDIT to larger datasets to explore further its capacity, and other vision applications, potentially enhancing generalization and interpretability in complex visual recognition tasks.